# DRCD: a Chinese Machine Reading Comprehension Dataset

**Chih Chieh Shao** and **Trois Liu** and **Yuting Lai** and **Yiying Tseng** and **Sam Tsai**
{cchieh.shao,trois.liu,yuting.lai,yiying.tz,i-sam.tsai}@deltaww.com
Delta Research Center
Delta Electronics, Inc.

## Abstract

In this paper, we introduce DRCD (Delta Reading Comprehension Dataset), an open domain traditional Chinese machine reading comprehension (MRC) dataset. This dataset aimed to be a standard Chinese machine reading comprehension dataset, which can be a source dataset in transfer learning. The dataset contains 10,014 paragraphs from 2,108 Wikipedia articles and 30,000+ questions generated by annotators. We build a baseline model that achieves an F1 score of 89.59%. F1 score of Human performance is 93.30%. The dataset is available at https://github.com/DRCKnowledgeTeam/DRCD.

## 1 Introduction

Machine reading comprehension (MRC) is the task of understanding paragraphs, and integrating it with what the human reader already knows. MRC systems can read documents written for humans and answer questions about the contents of such documents. In today's business world, all parties expect fast response times and easy access to information. For example, customers are often impatient to obtain answers to critical questions about products before purchasing. Novice workers look for timely support from experienced staff to help them solve problems. In the above cases, MRC can fill in for human experts and answer most questions immediately.

Although rule-based approaches to MRC have been developed and applied, the high labor cost of rule maintenance and the difficulty handling variations on the same questions limit the applications of rule-based MRC. Machine-learning-based approaches can mitigate these two problems. However, they require large enough datasets for training of MRC models.

> ……其他一些人，如伊本·西那醫師，則較為重視硫酸的種類以及它們在醫學上的價值……
> 在 1746 年，John Roebuck 則運用這個原則，開創鉛室法，以更低成本有效地大量生產硫酸。……
>
> ... Others, such as Ibn Sīnā, are more concerned about the type of sulfuric acid and their medical value.
> In 1746, John Roebuck applied this method to create lead chamber process to efficiently produce large quantities of sulfuric acid at a lower cost.
> **Question:** 重視硫酸的種類以及它們在醫學上的價值地為哪位醫師？
> Who value the type of sulfuric acid and their medical value?
> **Answer:** 伊本·西那 / Ibn Sīnā
> **Question:** 鉛室法於西元幾年開創？ / When did lead chamber process being invented?
> **Answer:** 1746 年 / 1746

Figure 1: Example from DRC Dataset.

Thanks to advances in and widespread adoption of deep learning and natural language processing, several large-scale MRC datasets have been compiled (Lai et al. 2017; Hermann et al. 2015; Cui et al. 2016; Rajpurkar et al. 2016; Nguyen et al. 2016; He et al. 2017), providing sufficient training data for deep learning MRC. These datasets have different research purposes and different task definitions, but they can be classified into four types by answer types: multiple choice, cloze-style, span-based, and user-log, we will describe more in Session 2. Multiple-choice datasets provide answer candidates and answers of cloze-style datasets were constrained to be a single word. Therefore, both of them are inappropriate for search scenario. Span-based and user-log datasets are more suited to our goal. There are two main abilities required in a search scenario, search, and comprehension. Models apply to user-log datasets need to learn both of the abilities. However, the state-of-the-art results of user-log datasets (e.g. MSMARCO) are still far behind human performance. It still remains a challenging problem in research. Furthermore, we want to test the ability of our model to comprehend text separately instead of learning both abilities at the same time. Therefore, we choose span-based datasets as our research target.

Most of existing MRC datasets are created from English corpus, and part of the others are generated from simplified Chinese corpus. However, to the best of our knowledge, no large-scale traditional Chinese MRC dataset has been compiled yet. In this paper, we introduce DRCD, an open domain MRC dataset, consisting of 10,014 paragraphs from 2,108 Wikipedia articles and 33,941 question-answer pairs.

## 2 Related Work

Recently, many MRC datasets have been constructed for different tasks and with different methods. Here we describe the four types of MRC dataset and give examples of each.

**Multiple-choice:** Multiple-choice MRC formulates the MRC task as an option selection problem. Multiple-choice datasets can easily be adapted from school examinations without much human labeling. Many previous works like Khashabi et al. (2016), Shibuki et al. (2014), Penas et al. (2014), Rodrigo et al. (2015) compiled MRC datasets from various levels of multiple-choice tests. Richardson et al. (2013) created MCTest, which contains 660 stories, 2,640 questions (4 per story) and 10,560 answer choices (4 per question) designed for 7-year-old children. Lai et al. (2017) constructed RACE, which contains 27,933 passages and 97,687 questions written for middle and high school students from 12–18 years old. But Lai et al. (2017) also indicate that multiple-choice MRC datasets are often far from sufficient for the training of advanced data-driven MRC models because of the expensive data-generation process by human experts.

**Cloze-style:** Cloze-style MRC formulates the task as the prediction of missing words in a sentence. Since cloze-style datasets can be constructed without human labeling, it is more practicable to compile one large enough for a data-demanding approach like deep learning. In English, Hermann et al. (2015) created a corpus from CNN and Daily Mail news summaries, and Hill et al. (2015) built the Children's Book Test. In Chinese, Cui et al. (2016) constructed a cloze dataset from the People's Daily news articles and another consisting of children's fairy tales. Though many deep learning models have been applied to these public datasets with impressive results, Chen et al. (2016) showed that cloze-style datasets require less reasoning and inference than previously assumed. This may be because the answers to cloze questions are single words or entities, which are relatively easier to guess than the answers to span-based dataset.

**Span-based dataset:** Span-based MRC assumes that the answer to each question can be found in the reference document. Rajpurkar et al. (2016) constructed the first span-based dataset, SQuAD, which has over 100,000 questions. Joshi et al. (2017) proposed TriviaQA, which includes 95,000 question-answer pairs from 14 trivia and quiz league websites. According to their analysis, TriviaQA has relatively complex, compositional questions compared with other large-scale datasets. Another difference between SQuAD and TriviaQA is that each question in SQuAD refers to only one evidence document, while questions in TriviaQA refer to multiple documents.

**User log dataset:** User log datasets are constructed from real-world search logs. Nguyen et al. (2016) released the MSMARCO dataset with 100,000 queries and answers. In MSMARCO, all questions are real anonymized user queries from the Bing search engine, and the evidence documents used as context passages are real web documents in Bing's index. He et al. (2017) constructed a Chinese user log dataset, DuReader, from the Baidu search engine and Baidu Zhidao, a question answering community site. Their dataset contains 200,000 questions, over 1 million documents, and over 420,000 human-generated answers.

## 3 Dataset Collection

In this work, we follow the method proposed by Rajpurkar et al. (2016) to collect Wikipedia data in three stages: passage curation, question-answer collection, and additional answer selection. An example is shown in Table.1.

**Passage curation.** To filter more informative entries, we calculate the top 10,000 articles of Chinese Wikipedia [1] using Project Nayuki's Wikipedia's internal PageRank. These entries were randomly assigned to annotators, and the annotators were asked to extract paragraphs from Wikipedia pages.

To further clean the data, annotators took article literal content only, ignoring images, figures, tables, bullet point, and links. All parentheses, square brackets with their contents have also been removed, which makes the article more fluent and does not affect its meaning. After cleaning, we choose the paragraphs between 250 and 1,500 Chinese characters, containing 10,014 paragraphs from 2,108 Wikipedia articles.

**Question-answer collection.**

In this stage, annotators were asked to read the whole article then generate at least 3 to 5 questions. The answer to each question must contain in the paragraph. The other criteria is that annotators are not allowed to copy the sentence in the paragraph as the question directly, they need to ask the question in their own way.

Additionally, we encourage annotators to ask the hard question that the answer can be a description or a sentence in the paragraph, which are relatively harder than questions for entities.

We also ask annotators to provide the specific question, to prevent from multiple answer situation. For example, if one asks: "Where did Obama live?" The paragraph might contain multiple answers like Hawaii or Washington, D.C. We suggest that annotators can add more information like date in the question to specify the answer. For example, "Where did Obama live in 2018?" will be better.

In order to verify whether the model learned how to inference or just doing task like named entity recognition, we ask annotators to annotate the exact location in case the answer appears multiple times in the paragraph.

---

[1] The Chinese Wikipedia dump is obtained in the data 2017/03/20

| Question Type | Percent (%) | Example keywords |
|---|---|---|
| how | 5.30 | 如何 |
| what | 28.42 | 什麼 |
| when | 13.59 | 何時 |
| where | 4.98 | 哪裡 |
| which | 30.96 | 何種 |
| who | 10.46 | 誰 |
| why | 0.27 | 為何 |
| other | 5.97 | X |

Table 2: Question types of DRCD.

We collect 26,932 questions in 8,014 paragraphs as training set, 3,524 questions in 1,000 paragraphs as development set, and 3,485 questions in 1,000 paragraphs as test set.

**Additional answer collection.** To evaluate human performance on our dataset. We obtained 1 additional answers to each question in development set and test set. Annotator was shown only questions and paragraphs and asked to select the shortest span in the paragraph that answer the question.

## 4 Dataset Analysis

We analyze training set and development set of the dataset, including paragraph length, question type, answer type, and the difficulty of DRCD.

**Question type:** To investigate the distribution of question types in DRCD. First, we sampled 660 questions randomly and then we construct the keyword list of each question types. Finally, we classify all questions into 7 types according to the keyword list. See Table 2 for the distribution and keyword sample of each question.

**Answer type:** We categorize the answers automatically into three types, numeric, entity, and description. First, we separate description and non-description answers. If the length of the answer is larger than 6 characters, the answer will be categorized as description type. Then, we further split non-description answers into the numeric and non-numeric group. The answer contains only numbers, Chinese numbers, and Chinese measure words will be categorized as a numeric type. Remainder are categorized as entity type. According to the statistics, the proportion of the answer types is 18.03% of numeric type, 70.45% of entity, and 11.50% of description.

**Statistics on Length:** On average, the length of paragraphs, questions, and answers are 435.8, 21.07 and 4.86 Chinese character respectively. We also examine SQuAD dataset. The average document length is 116.63 words and the average length of questions and answers are 10.06 and 3.16. Paragraphs in DRCD are much longer than previous dataset due to the reason that we ask annotator to separate paragraph by the topic of the paragraph instead of automated separate by period.

## 5 Experiment

In this section, we implement MRC systems with four different models compared with TF-IDF method. F1 score and exact match from Rajpurkar et al. (2016) are used as the evaluation metrics. Both metrics ignore punctuations. In F1 score metric, we consider predictions and ground truth as bag of Chinese character.

**Human Performance:** We assess human performance on development set and test set of DRCD. For each paragraph in development set and test set, we involve another annotator different from the one that constructs the paragraph and question-answer pair to answer the question and treat the answer as human prediction. The resulting human performance score on the test set is 80.43% for exact match metric, and 93.30% for F1.

**Baseline Systems:** We implement one basic method and four state-of-the-art models as the baseline. we use EternalFeather project [2] to process wikidump data [3], translate the text into traditional Chinese using OpenCC [4] . We use EternalFeather project to train a CBow word embedding provided by Mikolov et al. (2013) with 300 dimensions on the processed wiki text. We use this pre-trained word embedding in R-Net, QANet and BiDAF model.We use TF-IDF as basic traditional method.We count TF-IDF score of every sentence in paragraph and question. For each question, we find the most similar sentence in the related paragraph using cosine similarity and consider it as answer. The TF-IDF method achieve F1 score 17% but get 0.05% on exact match score. Wang et al. (2017) proposed R-net, which is a widely used MRC model. We adjust Yereval Project[5] to process Chinese character and achieve F1 score 38% and exact match score 23.8%. We use BiDAF which is implemented by He et al. (2017). We use pre-trained word embedding instead of randomly initialize. The other hyper-parameters remain the same as He et al. (2017). The result of BiDAF model is F1 score 51.18% and exact match score 28.08%. Adams et al. (2018) propose QANet. We adapt non-official implementation[6] and we change word tokenizer to jieba[7] in order to tokenize Chinese word. We get F1 score 78.03% and exact match score 65.56% without changing the hyperparameter setting of QANet model. BERT is released by Devlin et al. (2018). We use the pre-trained Chinese representation model[8] and fine-tune on DRCD using released code[9]. The final result of BERT model is F1 score 89.59% and Exact Match score 82.34%.

## 6 Conclusion and Future Work

We introduce DRCD, a new MRC dataset, which is first large-scale reading comprehension dataset in traditional Chinese. The dataset contains 10,014 paragraphs and 40,410 question-answer pairs from 2,108 Wikipedia articles. We aim to use this dataset to be the source dataset in transfer learning.

The deep learning method (R-Net, BiDAF, QANet and BERT) all have better performance than the traditional method. The result show that without using the structured annotation, the traditional method, such as TFIDF cannot achieve good performance on DRCD. Insight to the result of deep learning model, BERT get the outstanding performance beyond these models with F1 score 89.59% and Exact Match score 82.34%. But the F1 score of BERT is still lower than the human performance. This means that DRCD is still quite complicated.

---

[2] https://github.com/EternalFeather/Word2Vec-on-Wikipedia-Corpus
[3] zhwiki-20180320-pages-articles.xml.bz2
[4] https://github.com/BYVoid/OpenCC
[5] https://github.com/YerevaNN/R-NET-in-Keras
[6] https://github.com/NLPLearn/QANet
[7] https://github.com/fxsjy/jieba
[8] https://storage.googleapis.com/bert_models/2018_11_03/chinese_L-12_H-768_A-12.zip
[9] https://github.com/google-research/bert

In future work, we will focus on industrial data, which is our goal field to make next-generation search engine and question answering system. We expect we can improve annotation process and further adjust our task based on feedback from the community. We hope this dataset can promote the research in traditional Chinese reading comprehension. Our long-term goal is to apply different techniques in the applications that can be used in industry. Machine reading comprehension for search engine and question answering system is our first target.

## Acknowledgments

We are immensely grateful to Richard Tzong-Han Tsai, professor of National Central University, for his comments that greatly improved the manuscript. We would also like to show our gratitude to Hung-Yi Lee, assistant professor of National Taiwan University, for sharing his pearls of wisdom with us during the course of this research.